\documentclass[letterpaper, 10 pt, conference]{ieeeconf}  

\IEEEoverridecommandlockouts                              

\usepackage{graphicx}
\usepackage{amsmath,amssymb,amsfonts}
\usepackage{cite}
\makeatletter
\let\NAT@parse\undefined
\makeatother
\usepackage{hyperref}

\title{\LARGE \bf
Differential Barometric Altimetry for Submeter Vertical Localization and Floor Recognition Indoors
}

\author{
    Yuhang Zhang$^{1}$ and S\"{o}ren Schwertfeger$^{1,*}$
    \thanks{*Corresponding author.}%
    \thanks{$^{1}$All authors are with the Key Laboratory of Intelligent Perception and Human-Machine Collaboration – ShanghaiTech University, Ministry of Education, China.
    {\tt\small \{zhangyh12024, soerensch\}@shanghaitech.edu.cn}}%
}

\begin{document}

\maketitle
\thispagestyle{empty}
\pagestyle{empty}

\begin{abstract}

Accurate altitude estimation and reliable floor recognition are critical for mobile robot localization and navigation within complex multi-storey environments. In this paper, we present a robust, low-cost vertical estimation framework leveraging differential barometric sensing integrated within a fully ROS-compliant software package. Our system simultaneously publishes real-time altitude data from both a stationary base station and a mobile sensor, enabling precise and drift-free vertical localization. Empirical evaluations conducted in challenging scenarios—such as fully enclosed stairwells and elevators, demonstrate that our proposed barometric pipeline achieves sub-meter vertical accuracy (RMSE: $0.29\,\mathrm{m}$) and perfect ($100\,\%$) floor-level identification. In contrast, our results confirm that standalone height estimates, obtained solely from visual- or LiDAR-based SLAM odometry, are insufficient for reliable vertical localization. The proposed ROS-compatible barometric module thus provides a practical and cost-effective solution for robust vertical awareness in real-world robotic deployments. The implementation of our method are released as open source at \url{https://github.com/witsir/differential-barometric}.

\end{abstract}

\section{Introduction}
\subsection{Motivation: Vertical Awareness in Indoor Robotics}

Accurate altitude estimation and unambiguous floor recognition are fundamental prerequisites for reliable localization and navigation of mobile robots in complex indoor environments. As buildings are intrinsically three-dimensional (they comprise stacked floors, elevators, staircases, and ramps), purely planar $(x,y)$ navigation is inadequate~\cite{jung2024munes,wang2017uneven}. Without precise knowledge of their vertical state, i.e., which floor they currently occupy, robots may load an incorrect map layer, triggering severe localization errors and subsequent mission failures.

Contemporary robotic‑navigation frameworks (e.g., ROS stacks) therefore maintain separate 2‑D occupancy grids or semantic floor maps for each story and require robots to identify their present floor online to enable correct map switching~\cite{palacin2023procedure}. Even sophisticated multi‑floor representations, such as hierarchical or voxelized 3‑D maps, still rely on accurate altitude cues to detect floor transitions and preserve global consistency~\cite{karg2010consistent,jung2024munes,wei2021ground,shan2019rgbd}. Likewise, SLAM systems that employ planar constraints must incorporate explicit vertical‑awareness mechanisms; otherwise, unnoticed height transitions (e.g., via stairs or elevators) manifest as map misalignments and accumulated drift~\cite{wei2021ground,campo2020orbslam3}. In general, robot localization, especially when dealing with the kidnapped robot problem, welcomes clues/ initial guesses to improve computation speed and accuracy \cite{xie2023robust}.

Empirical studies confirm that inaccuracies in altitude estimation and discrete floor detection during inter‑floor motions routinely degrade localisation quality and corrupt map integrity, leading to mission‑critical breakdowns in service and logistics deployments~\cite{jung2024munes,wang2017uneven}. Consequently, vertical awareness is not a mere enhancement but a core capability for dependable indoor robot operation.

\subsection{Barometric and Barometer-Augmented Methods for Indoor Vertical Estimation}\label{sec:approaches}

Robust vertical localization requires mechanisms that bound drift, provide absolute references, and leverage the low‑cost barometric cues that many state‑of‑the‑art systems rely on for their final accuracy boost. In this section, we categorize existing vertical estimation methods into four representative approaches.

\subsubsection*{1) Barometric Altimetry and Floor Assignment}
Single‐sensor barometers convert pressure to height but suffer from HVAC (Heating, Ventilation and Air Conditioning) drift and handset offsets~\cite{li2013baro, tsubouchi2023faneffects}. Robust variants exploit reference barometers—either dual nodes~\cite{bao2024uwbbaro} or crowdsourced bases~\cite{shen2015barfi}—and probabilistic floor models~\cite{fetzer2023baroindoors,xia2015multiplebaro}.  Reported root mean squared (RMS) errors fall below $0.3\,\mathrm{m}$ after calibration, establishing barometry as the lightest infrastructure for vertical priors.

\subsubsection*{2) Inertial \& Vision–Aided Prediction}
IMU double integration alone diverges quickly; mode‑specific resets (e.g., stairs or elevator detection) can limit drift to a few meters over typical vertical traverses~\cite{hager2025barometer}. Fusing a barometer in an EKF or factor graph framework further constrains vertical drift to the centimeter scale~\cite{sabatini2014sensorfusion}. Direct optical flow combined with IMU can also achieve low-bias altitude estimation in short-range indoor flights without requiring external priors, yielding RMS errors around $2.5\,\mathrm{cm}$~\cite{chirarattananon2018direct}. For long-range or multi-floor settings, monocular VIO systems (e.g., \textsc{VINS‑Mono}~\cite{qin2018vinsmono}) can refine visual scale, especially when fused with barometric measurements.

\subsubsection*{3) RF‑Assisted Height Localisation}
UWB 3D localization offers centimeter accuracy in line‑of‑sight; coupling its Z output with a barometer absorbs slow atmospheric drift~\cite{bao2024uwbbaro}.  Wi‑Fi/BLE fingerprinting meets the FCC E911\footnote{Refers to the FCC E911 Vertical Location Accuracy Mandate, which requires sub-3 meter RMS error in 80\% of cases for Z-axis localization.} vertical mandate in open halls, yet its performance degrades in stairwells; adding barometric clusters (BarFi) boosts floor recognition to $> 95\%$ across heterogeneous devices~\cite{shen2015barfi,mundlamuri2025sdr-fi-z}. These RF or RF–baro hybrids can achieve an accuracy that is sufficient for many practical robotic applications.

\subsubsection*{4) Geometric SLAM with Altitude Constraints}
\label{sec:geo_slam}
Vision-/LiDAR-based SLAM supplies metric structure but drifts along gravity, especially in texture-poor shafts. Injecting ceiling/floor plane factors~\cite{wei2021ground} or discrete floor priors~\cite{karg2010consistent} stabilizes the $Z$ axis without costly loop closures. Furthermore, recent work by Dubois \textit{et al.} (2024) demonstrates a novel approach to tightly integrate barometric altitude constraints within ICP-based SLAM back-end, reducing vertical drift by approximately $84\,\%$ in under-constrained environments~\cite{dubois2025pressureICP}. It provides the first credible evidence that barometer–SLAM coupling can significantly stabilize $Z$-axis drift when fused tightly in the optimization backend.

\begin{figure}[t]
\centering
\includegraphics[width=0.80\linewidth, trim=0 250 320 0, clip]{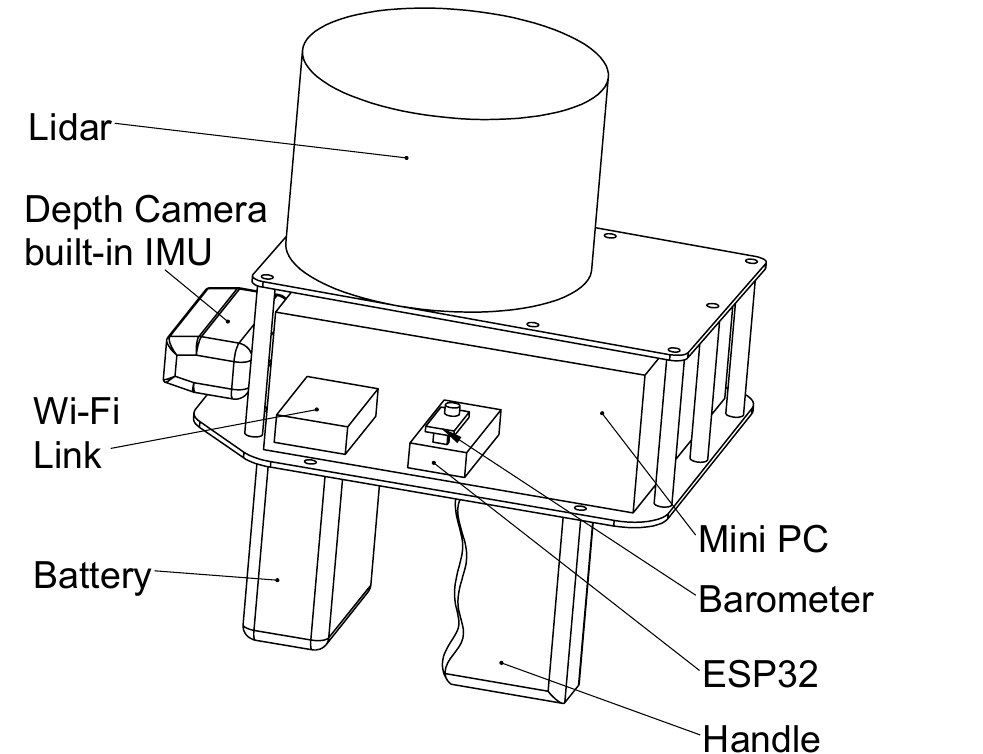}
\caption{The custom-designed SLAM device used in our experiments.}
\label{fig:SLAMdevice}
\end{figure}

\subsection{Limitations of Existing Vertical Estimation Approaches}
Despite notable progress, current indoor height–estimation techniques still face several roadblocks that map closely—but not perfectly—to the taxonomy in Sec.~\ref{sec:approaches}.

\subsubsection*{1) Infrastructure dependence and limited robustness in vertical conduits}  
Radio‑frequency approaches—including UWB, BLE beacons, and Wi‑Fi RSS/CSI fingerprinting—rely on pre‑installed anchors or a sufficiently dense access‑point (AP) grid. Wi-Fi fingerprinting requires an extensive site survey to associate RSSI values with ground-truth locations. The system requires an offline phase to build a fingerprint database. Sparse AP layouts or ad‑hoc deployments inflate cost and often leave stairwells and elevator shafts with inadequate coverage. Even when infrastructure is present, severe multipath and signal occlusion in these confined spaces degrade Z‑accuracy to an unverified level \cite{bao2024enhanced,bao2024uwbbaro,marjasz2024comprehensive,mundlamuri2025sdr-fi-z}. 

\subsubsection*{2) Visual/LiDAR SLAM under‑validated in stairwells and elevator cars}  
While vision‑ and LiDAR‑based SLAM demonstrate sub‑decimetre performance in open corridors, few studies report systematic results in narrow, repetitive, or poorly lit vertical passages.  Comparative evaluations show pronounced drift for monocular methods in texture‑poor shafts and for LiDAR in long glass corridors lacking loop closures \cite{zhao2024comprehensive,cong2024efficient,wei2021ground}.  Rigorous, public datasets covering such scenarios remain scarce.

\subsubsection*{3) Absence of Networked Base-Station Barometer Solutions}  
Some commercial sensors (e.g., MEMS IMUs) may embed a barometer, yet the available software stacks only expose local pressure readings. The crucial step of linking a mobile barometer to a remote reference barometer so that differential pressure yields an absolute altitude is rarely supported out-of-the-box. Existing open-source examples offer fragmentary drivers or platform-specific scripts, but no lightweight, cross-platform framework that simultaneously publishes base-station pressure and mobile pressure through a uniform ROS interface.  Consequently, practitioners must devise their own hardware and communication layer before any field deployment or fair comparison can be attempted \cite{li2013baro,fetzer2023baroindoors,sabatini2014sensorfusion}.

\subsubsection*{4) Fragmented evaluation protocols and missing cross‑modality benchmarks}  
Many studies assess a single modality under limited conditions, hindering objective comparison across sensors in identical vertical scenarios (e.g., elevators, emergency stair cores). Stairwells and elevators constitute the two primary settings in which indoor robots must perceive and negotiate vertical motion. A complete evaluation of any vertical-sensing pipeline should therefore measure a joint error across both scenarios. The absence of standardized datasets, metrics, and plug‑in evaluation scripts impedes reproducible benchmarking and slows adoption.

Overall, these gaps motivate a vertically aware, infrastructure‑light framework—backed code that can be readily integrated into existing robotic systems and rigorously evaluated in challenging vertical environments.

\subsection{Our Contributions}

To address the above limitations, this paper makes the following contributions:

\begin{itemize}
    \item We design and implement a ROS plug-and-play barometer/Wi-Fi height estimation module. It consists of a low-cost hardware stack (a barometric sensor and an Arduino-compatible board) and a ROS node that publishes height estimates as standard ROS topics. These components are integrated into a small SLAM unit as shown in Figure~\ref{fig:SLAMdevice}. Additionally, by connecting to a base pressure observation station via the internet, the system can utilize the known base altitude to provide altitude information for robot navigation and localization tasks, such as in scenarios where the robot's session is intermittent or the robot is in a kidnapped situation. This module can be integrated into any existing ROS system without code refactoring, directly addressing reproducibility and integration gaps.
    \item We conduct a series of empirical evaluations on Vision-based SLAM, LiDAR-based SLAM, Wi-Fi fingerprinting, and barometer-based methods in a fully enclosed four-story stairwell and an elevator shaft, offering a side-by-side quantitative comparison under controlled conditions for height estimation and floor detection. We built a hand rig SLAM equipment as shown in Figure~\ref{fig:SLAMdevice} to manage all these experiments.
\end{itemize}

\section{Related Work}
\label{sec:related}

Research on vertical localization with the barometer for indoor robots can be grouped into three threads:  
\textit{(i)}~single-barometer height-change detection,  
\textit{(ii)}~differential barometry and multi-sensor fusion, and  
\textit{(iii)}~robot-centric frameworks that evaluate barometers in elevators and stairwells.  
We review representative robotics literature and highlight the gap that our proposed plug-and-play, base-station-referenced barometer solution is designed to fill.

\subsection{Single-barometer Floor-Transition Detection}

Early ground robots mounted a single pressure sensor to segment multi-storey maps.  
Özkil \textit{et al.} used pressure jumps to initialize a new 2D map layer whenever their hospital robot changed floors~\cite{ozkil2011barometric}.  
Li \textit{et al.} reported meter-level RMS errors with a lone barometer but showed that relative pressure still sufficed to trigger discrete floor labels~\cite{li2013baro}.  
More recently, Jung \textit{et al.} embedded a barometer in the \textsc{MuNES} planner to recognize stairs and elevators during multi-floor LiDAR navigation, yet altitude remained relative to the robot start pose~\cite{jung2024munes}.  
These works confirm that on-board barometers can effectively detect vertical transitions, but none deliver an absolute prior height at robot initialization time.

\subsection{Differential Barometry and Fusion}

Differential approaches reference a stationary barometer to cancel atmospheric drift.  
Bao \textit{et al.} fused dual barometers with UWB, attaining $\pm 5\,\mathrm{cm}$ vertical RMS in line-of-sight labs~\cite{bao2024uwbbaro}.  
Fetzer \textit{et al.} combined relative pressure, Wi-Fi, and IMU in a particle filter and demonstrated $<0.3\,\mathrm{m}$ floor-assignment error across three buildings—yet their code targets Android, not ROS~\cite{fetzer2023baroindoors}.  
Hager \textit{et al.} integrated a base–mobile barometer pair with an IMU classifier to distinguish elevators, escalators, and stairs in an indoor navigation stack~\cite{hager2025barometer}.  
Smartphone systems such as \textsc{FloorPair}~\cite{yi2019floorpair} and BarFi~\cite{shen2015barfi} reach $>95\,\%$ floor accuracy via differential pressure maps, but robotic evaluations remain absent.

\subsection{Robot-centric Evaluations in Elevators and Stairwells}

Elevators and fire-stairs are notoriously challenging for LiDAR, vision, and RF localization due to sparse features and signal occlusion.  
Jung \textit{et al.} benchmarked LiDAR SLAM in elevator rides but did not quantify vertical error~\cite{jung2024munes}.  
Hager \textit{et al.} reported barometer–IMU performance inside elevators but released no ROS package~\cite{hager2025barometer}.  
To our knowledge, no prior study provides a ROS-ready node that (i) streams drift-free altitude referenced to a base barometer from system startup, and (ii) publishes discrete floor labels, while being validated in both elevators and enclosed stairwells.

\subsection{Summary}

Existing robotics literature verifies that barometers excel at detecting floor transitions, and differential barometry can yield sub-decimeter accuracy, but implementations are either phone-centric or require manual calibration.  
No open-source, turn-key ROS solution currently delivers base-referenced altitude plus floor IDs.  
Our work fills this void by releasing a plug-and-play ROS module that communicates with a networked base barometer, providing absolute height at boot time and achieving sub-meter accuracy in the two vertical conduits most critical for indoor robots: elevators and stairwells.

\section{Method}
\label{sec:method}

\begin{figure}[t]
\centering
\includegraphics[width=0.90\linewidth, trim=0 300 392 0, clip]{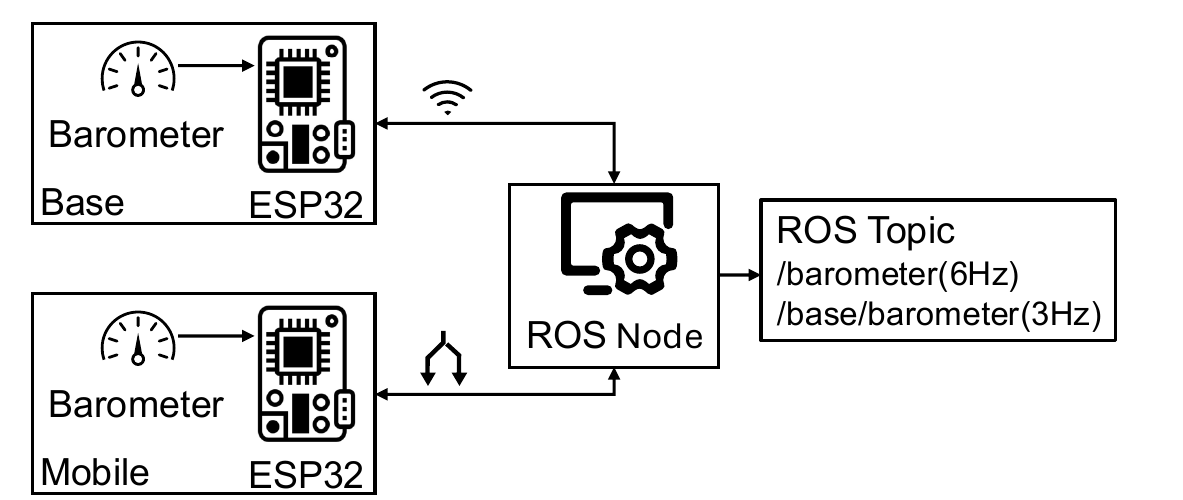}
\caption{The data pipeline of the differential barometer ROS package.}
\label{fig:barometer_ros_pipeline}
\end{figure}

Figure~\ref{fig:barometer_ros_pipeline} illustrates how a ROS node connects to two barometers and publishes their data.
\subsection{Pressure-to-Height Map (ISA Troposphere)}
For both the base and mobile barometers, we convert pressure to geometric height using the International Standard Atmosphere (ISA) tropospheric model:
\begin{equation}
  h(P,T) = \frac{T}{L}\left[1 - \left(\frac{P}{P_{0}}\right)^{\kappa} \right],
  \qquad
  \kappa \triangleq \frac{R L}{g} \approx 0.1903,
  \label{eq:isa-abs}
\end{equation}
where $T$ is absolute temperature, $L = 0.0065\,\mathrm{K\,m^{-1}}$, $R = 287.05\,\mathrm{J\,kg^{-1}\,K^{-1}}$, $g = 9.80665\,\mathrm{m\,s^{-2}}$, and $P_0$ is the current sea-level reference pressure retrieved from a public meteorological API.

Applying~\eqref{eq:isa-abs} to the base ($b$) and mobile ($m$) sensors yields
\begin{equation}
  h_b = f(P_b, T_b),\qquad h_m = f(P_m, T_m),
\end{equation}
with $f(\cdot) \equiv h(\cdot)$ as defined in~\eqref{eq:isa-abs}.
Because both heights share the same $P_0$, subtracting them cancels absolute-pressure bias and yields the \emph{relative} altitude of the robot with respect to the ground floor:
\begin{equation}
  \Delta h = h_m - h_b.
  \label{eq:delta-h}
\end{equation}

\subsection{Relative Calibration of Multiple Barometers}
When $N$ barometers are collocated on the same horizontal plane, the ideal readings would coincide up to noise. In practice, manufacturing tolerances introduce sensor-specific constant offsets. We therefore perform an offset-only, relative calibration using two days of collocated data.

\subsubsection*{1)Preprocessing and Time Alignment}
Let $\{p_i(t), T_i(t)\}$ denote the raw pressure and temperature from sensor $i\in\{1,\dots,N\}$ at time $t$. We construct a uniform grid $\mathcal{T}=\{t_1,\dots,t_M\}$ by resampling each data stream to $\Delta = 30\,\mathrm{s}$ with averaging, then perform an inner join across sensors. To suppress impulsive artifacts, we discard timestamps that violate practical jump thresholds:
\begin{multline}
\mathcal{T}' = \Big\{ t_m\in\mathcal{T} :
\; |p_i(t_m)-p_i(t_{m-1})|\le 1\,\mathrm{hPa},\\
|T_i(t_m)-T_i(t_{m-1})|\le 1^{\circ}\mathrm{C},\; \forall i \Big\},
\end{multline}
which yields $M'=\lvert\mathcal{T}'\rvert$ aligned samples.

\subsubsection*{2)Offset Model and Estimation}
With all sensors sharing the same ambient conditions while collocated, we model
\begin{equation}
  \begin{aligned}
    p_i(t) &= p^{\star}(t) + \beta_i^{(p)} + \varepsilon_i^{(p)}(t),\\
    T_i(t) &= T^{\star}(t) + \beta_i^{(T)} + \varepsilon_i^{(T)}(t),
  \end{aligned}
  \qquad t\in\mathcal{T}',
  \label{eq:offset-model}
\end{equation}
where $p^{\star}(t)$ and $T^{\star}(t)$ are the common (time-varying) signals, $\beta_i^{(p)}$ and $\beta_i^{(T)}$ are time-invariant offsets, and $\varepsilon_i^{(\cdot)}(t)$ are zero-mean noises. For identifiability, we fix the gauge by enforcing
\begin{equation}
  \sum_{i=1}^N \beta_i^{(p)}=0,\qquad \sum_{i=1}^N \beta_i^{(T)}=0.
  \label{eq:gauge}
\end{equation}
We estimate $\{p^{\star}(t)\}$ and $\{\beta_i^{(p)}\}$ by constrained least squares:
\begin{equation}
  \min_{\{p^{\star}(t)\},\{\beta_i^{(p)}\}} \sum_{t\in\mathcal{T}'}\sum_{i=1}^N \big(p_i(t)-p^{\star}(t)-\beta_i^{(p)}\big)^2
  \;\; \text{s.t.}\; \eqref{eq:gauge},
\end{equation}
and analogously for temperature. The normal equations admit closed-form solutions:
\begin{equation}
  \hat p^{\star}(t) = \tfrac{1}{N}\sum_{i=1}^N p_i(t),\qquad
  \hat\beta_i^{(p)} = \tfrac{1}{M'}\sum_{t\in\mathcal{T}'}\big(p_i(t)-\hat p^{\star}(t)\big)
  \label{eq:pressure-solution}
\end{equation}
\begin{equation}
  \hat T^{\star}(t) = \tfrac{1}{N}\sum_{i=1}^N T_i(t),\qquad
  \hat\beta_i^{(T)} = \tfrac{1}{M'}\sum_{t\in\mathcal{T}'}\big(T_i(t)-\hat T^{\star}(t)\big)
  \label{eq:temperature-solution}
\end{equation}
Equations~\eqref{eq:pressure-solution}--\eqref{eq:temperature-solution} precisely (i) average across sensors per timestamp to form a reference and (ii) time-average each sensor’s deviation from that reference to obtain a constant offset.

\paragraph*{Calibrated Readings and Use in ISA}
During operation, we subtract the learned offsets to obtain calibrated sequences
\begin{equation}
  \tilde p_i(t)=p_i(t)-\hat\beta_i^{(p)},\qquad \tilde T_i(t)=T_i(t)-\hat\beta_i^{(T)}.
  \label{eq:apply-calib}
\end{equation}
All height computations substitute $(\tilde P,\tilde T)$ into~\eqref{eq:isa-abs}. In particular, for the base/mobile pair used in~\eqref{eq:delta-h},
\begin{equation}
  \Delta h = h\big(\tilde P_m,\tilde T_m\big) - h\big(\tilde P_b,\tilde T_b\big),
\end{equation}
which removes the common $P_0$ bias and the sensor-specific additive biases estimated in~\eqref{eq:pressure-solution}--\eqref{eq:temperature-solution}. For reporting, we denote
\begin{equation}
\begin{aligned}
  \texttt{pressure\_offset}_i &\triangleq \hat\beta_i^{(p)}\;[{\mathrm{hPa}}], \\
  \texttt{temperature\_offset}_i &\triangleq \hat\beta_i^{(T)}\;[^{\circ}\mathrm{C}].
\end{aligned}
\end{equation}

\subsection{Floor Indexing}
Given the measured floor heights $\{H_k\}$, we compute the floor index by nearest-neighbor assignment:
\begin{equation}
  \ell = \operatorname*{arg\,min}_{k} \big| \Delta h - H_k \big|.
\end{equation}

\subsection{Sensitivity to ISA Constants}
Using the default ISA constants $(L,R,g)$ perturbs the relative height estimate $\Delta h$ by less than $2\,\mathrm{cm}$ over our operating range—well below the sensor’s noise floor—so precise calibration of $(L,R,g)$ is unnecessary when the system outputs height differences. However, the calibration offsets in~\eqref{eq:pressure-solution}--\eqref{eq:temperature-solution} may drift over time due to sensor aging or wear, and should be periodically re-estimated to maintain accuracy.

\section{Experiments Setup}
\subsection{Hardware and Devices}

\begin{figure}[t]
\centering
\includegraphics[width=0.90\linewidth, trim=0 180 392 0, clip]{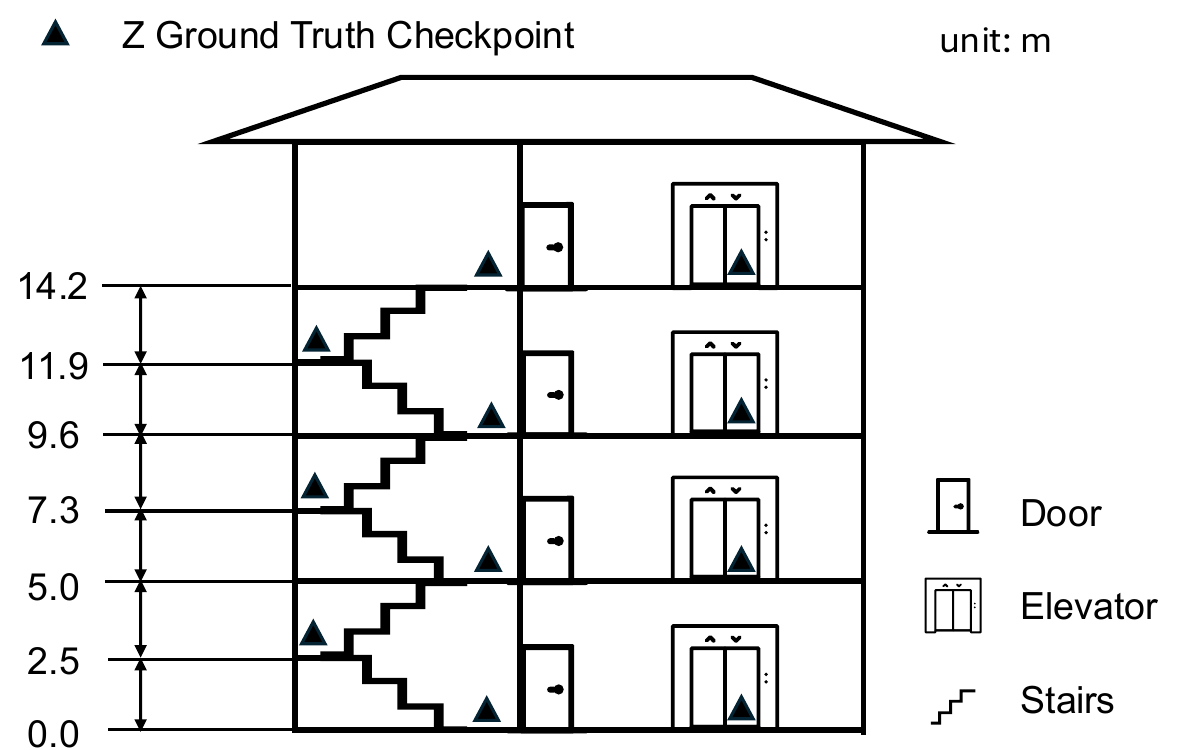}
\caption{Schematic of the experimental environment. The data collection trajectory covered a full ascent through the multi-story stairwell and a subsequent descent in the elevator.}
\label{fig:experiment_environment}
\end{figure}

We prepared a handheld SLAM rig, as shown in Figure~\ref{fig:SLAMdevice}. It integrates an Intel RealSense D435i RGB-D camera, which provides synchronized color and depth frames, alongside an on-board IMU for accelerometer and gyroscope measurements. A Velodyne VLP-16 LiDAR (16 channels) is mounted to acquire $10\,\mathrm{Hz}$ 3D point clouds. The rig is powered by a TB47S battery pack, and computation is handled by an ASUS PN53 ultra-compact mini PC. The system supports approximately three hours of continuous data recording. An ESP32 Devkit is connected via a serial cable to read a barometric sensor at about 6~Hz. An EDUP EP‑AC1681 adapter to serve as a Wi-Fi acquisition terminal. The rig also allows direct measurement of the vertical offset between the device and the device’s locate plane. We conducted physical measurements using a tape measure on stair landings and arrival platforms to obtain ground-truth heights, as shown in Figure~\ref{fig:experiment_environment}, denoted as Z Ground Truth Checkpoint.

\subsection{Environments and Datasets}
To thoroughly evaluate our proposed method under realistic conditions, we collected a custom dataset within a combined test area comprising a stairwell and an adjacent elevator hall in a multi-floor academic building, as shown in Figure~\ref{fig:experiment_environment}. The environment poses several significant challenges, such as inadequate illumination, limited Wi-Fi coverage, repetitive structural patterns, and low-texture surfaces, among others. In elevator cabins, SLAM systems frequently become ineffective due to the lack of observable relative motion between the sensor and the surrounding environment, which violates the fundamental assumptions of feature-based localization. As a result, it is necessary to rely exclusively on barometric and Wi-Fi-based methods to estimate vertical displacement.

\subsubsection*{1) Stairwell Data Collection.} Data collection commenced from a predefined initial point on the first floor, also known as checkpoint 1, proceeding up to the fourth floor via the stairwell. During this ascent, sensors recorded synchronized data streams, including monocular RGB images, depth images, LiDAR point clouds, barometric pressure(including base barometer), and Wi-Fi RSSI values. To acquire reliable ground-truth height measurements, we make sure we have the exact height above identified every checkpoint along the route, consisting of three inter-floor landings and four main floor platforms. Using direct physical measurements combined with known building geometry (floor-to-floor heights and inter-floor landing heights), we precisely calculated the true vertical positions at these reference points.

\subsubsection*{2) Elevator Data Collection.} Elevator sequences began from the next checkpoint after the stairs data collection on the fourth floor. Upon entering the elevator, we sequentially triggered floor-stop requests for each level, recording sensor data and ground-truth measurements at each stop. 

\subsubsection*{3) Wi-Fi Data Collection}
Vertical localization using Wi-Fi fingerprinting comprises two distinct phases: an offline fingerprint mapping phase and an online localization phase. First, a Wi-Fi fingerprint database is constructed by recording Received Signal Strength Indicator (RSSI) values at predefined reference points with known vertical positions along the experimental trajectory. Subsequently, during the online localization phase, the recorded RSSI patterns at unknown locations are matched against this database to estimate discrete vertical positions (e.g., specific floors). It should be noted that in this paper, Wi-Fi-based localization inherently provides discrete floor-level estimations rather than continuous altitude measurements, making it primarily for coarse floor identification rather than precise height tracking.

All sensor data streams—including monocular RGB images, depth measurements, point clouds, barometric readings, and Wi-Fi RSSI measurements—were recorded via ROS nodes and serialized into ROS bag files. This approach ensured precise synchronization of data and facilitated subsequent analysis.

\begin{figure}[t]
\centering
\includegraphics[width=0.80\linewidth, trim=0 0 0 0, clip]{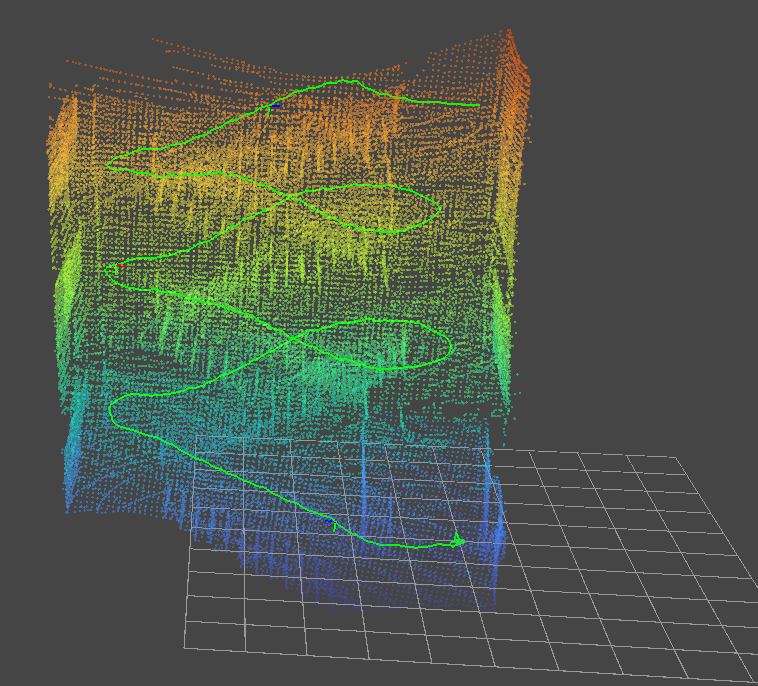}
\caption{3D point cloud with trajectory generated by GLIM.}
\label{fig:points_cloud}
\end{figure}

\subsection{Implementation Details}
We implemented a comprehensive sensing, communication, and processing pipeline to enable consistent data acquisition and accurate height estimation across all experimental scenarios. Below, we detail the barometric and Wi-Fi subsystems, visual and LiDAR SLAM-based height estimation methods, and discuss the platform-specific considerations.

\subsubsection*{1) Barometric Subsystem} We selected a low-cost HP206C barometer for both the stationary base node and mobile platforms. Two embedded software modules were developed: (i) the base node continuously broadcasts barometric readings containing timestamps, temperature, and pressure values over the wireless network; (ii) the mobile node transmits barometric measurements via serial communication. A dedicated ROS node on the mobile platforms converts these streams into standardized ROS topics (\texttt{/base/barometer} at 3Hz and \texttt{/barometer} at 6Hz), publishing temperature, pressure, and altitude.

\subsubsection*{2) Wi-Fi Subsystem} Wi-Fi RSSI data were collected using an EDUP EP-AC1681 adapter. Two specialized ROS nodes were developed: the first node collects initial RSSI measurements labeled with known positions to build a fingerprint map for indoor localization; the second node employs this fingerprint map for real-time vertical (z-axis) localization. Although full three-dimensional localization data are generated, we utilize only vertical height data for evaluating localization accuracy.

\subsubsection*{3) Visual and LiDAR Height Estimation} Due to differences in sensor cost, size, and practicality, visual and LiDAR-based approaches were assessed independently. For visual-based height estimation, we applied ORB-SLAM3~\cite{campo2020orbslam3} with RealSense D435i. LiDAR-based height estimation was separately performed using GLIM~\cite{koide2024glim}. Each estimation method’s height outputs were compared individually against ground truth data.

\subsection{Results and Analysis}

Table~\ref{tab:vertical_mse_floor} and Figure~\ref{fig:estimation_plot} summarize the vertical estimation performance of the four pipelines in both the enclosed stairwell and elevator scenarios. It should be emphasized that, in this study, SLAM is employed solely as a tool for height estimation; assessment of mapping accuracy and loop-closure performance is beyond the scope of this study.

\begin{table}[!h]
\caption{Comparison of vertical estimation error (RMSE) and floor detection accuracy}
\label{tab:vertical_mse_floor}
\centering
\begin{tabular}{lcc}
\hline
\textbf{Method} & \textbf{RMSE (m)} & \textbf{Floor Detection Accuracy (\%)} \\
\hline
\textbf{Barometer} & \textbf{0.291} & \textbf{100} \\
LiDAR-based SLAM & 3.480 & --\\
Visual-based SLAM & 3.718 & --\\
Wi-Fi & --$^{\mathrm{a}}$  & 81.8 \\
\hline
\multicolumn{3}{l}{$^{\mathrm{a}}$RMSE was not reported, as the Wi-Fi estimates are discrete values.}
\end{tabular}
\end{table}

\subsubsection*{1) Height Estimation Aided with Base Station Barometer}  
With a networked base barometer providing a reference pressure, the mobile barometer achieves an RMSE of $0.29\,\mathrm{m}$ over four storeys. Given that floor-to-floor\footnote{In the experiment, we also labeled checkpoints in the intermediate landing areas of all staircases. The minimum vertical distance between adjacent checkpoints is $2.3\,\mathrm{m}$.} spacing in the test building is $2.3\,\mathrm{m}$; this error is an order of magnitude smaller than the inter-floor distance, yielding $100\,\%$ correct-floor detection in both the stairwell and the fully enclosed elevator. The result corroborates prior reports of sub-decimetre drift after differential pressure compensation~\cite{xia2015multiplebaro,fetzer2023baroindoors} and demonstrates that a single low-cost sensor can satisfy the FCC E911 $3\,\mathrm{m}$ vertical mandate with a comfortable margin.

\subsubsection*{2) Visual and LiDAR SLAM Odometry}  
\begin{figure}[t]
\centering
\includegraphics[width=1.0\linewidth, trim=0 345 0 0, clip]{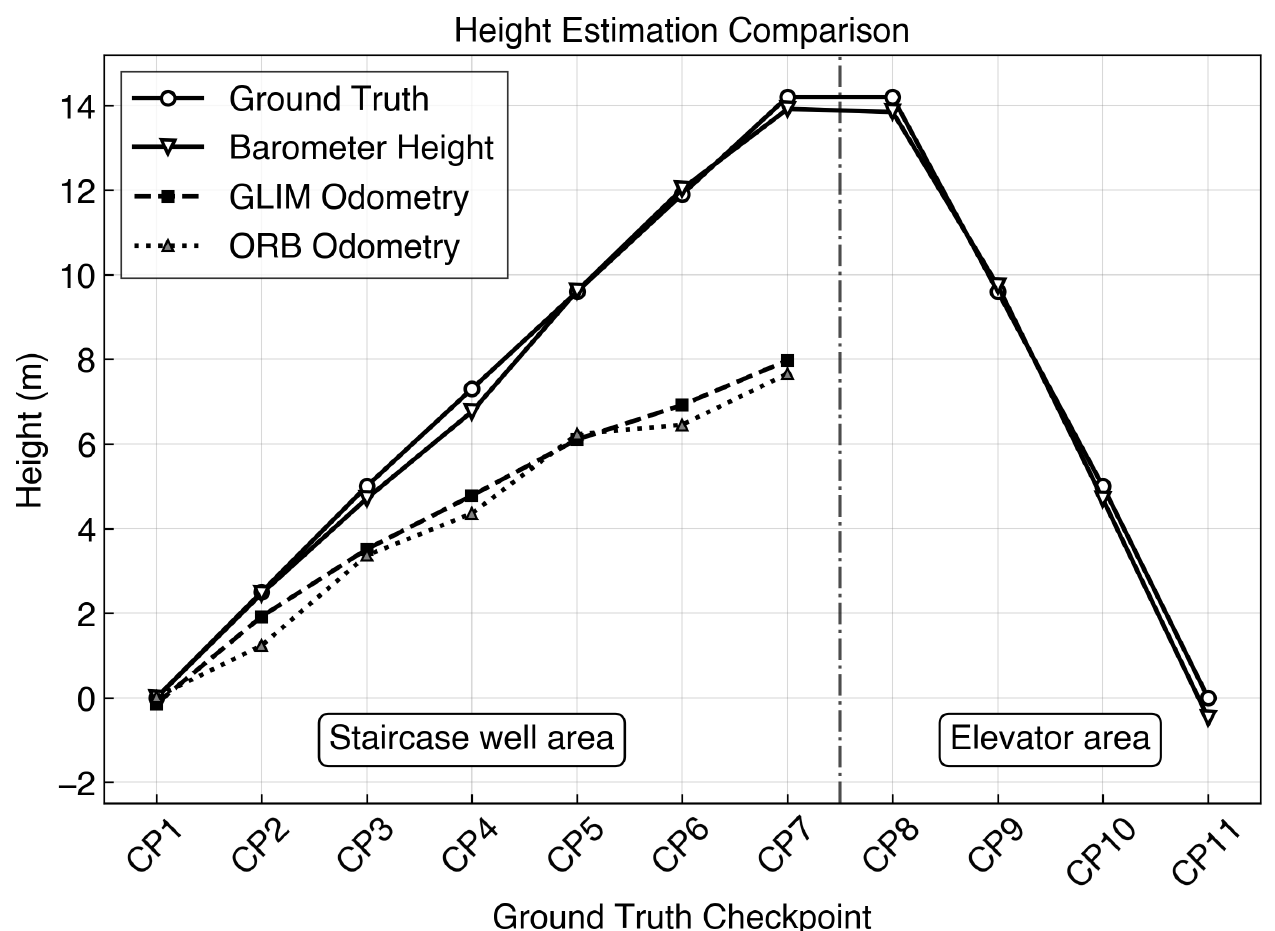}
\caption{Altitude estimation traces in stair (CP1–CP7) and elevator (CP7–CP11) scenarios.}

\label{fig:estimation_plot}
\end{figure}

ORB-SLAM3 (RGB-D + IMU) and GLIM (LiDAR + IMU) produce nearly overlapping altitude traces in the stairwell (Figure~\ref{fig:estimation_plot}), consistent with both systems relying on the RealSense inertial unit for $Z$ velocity integration. In the narrow stairwell, the device was moved along an almost helical path, which offered limited lateral parallax. We use GlIM to generate a 3D point cloud with a trajectory in the stair scenario as shown in Figure~\ref{fig:points_cloud}. Monocular depth scale (ORB-SLAM3) and ICP registration (GLIM) are therefore under-constrained along the gravity axis, so small IMU bias or time-synchronisation errors accumulate into metre-level offsets. Risers, handrails, and white walls repeat every half-landing, producing ambiguous feature matches for vision and nearly planar point clouds for LiDAR. Both conditions are known to cause pose-graph degeneracy and bias in the vertical component~\cite{zhao2024comprehensive,cong2024efficient}. Elevators represent typical SLAM-refused environments. Inside the enclosed cabin, sensors are disconnected from external geometry, preventing the tracking of relative motion.

\subsubsection*{3) Wi-Fi Fingerprinting}  
The fingerprint map, constructed under fully closed fire-door conditions, contains only eight distinct reference points. During tests, floor-level accuracy reaches $81.8\,\%$ where AP signals are heavily attenuated, especially in stairwells. The discrete-valued nature of RSSI matching precludes sub-floor resolution, confirming that Wi-Fi alone is insufficient for fine-grained altitude tracking in RF-shielded shafts.

\subsubsection*{4) Comparative Insight}  
The barometric solution outperforms all other modalities by more than an order of magnitude in RMSE while maintaining $100\,\%$ floor recognition. While valuable for horizontal odometry, SLAM-based methods are unreliable for inferring altitude in visually or geometrically degenerate spaces, and thus cannot function as a standalone solution for floor detection. Wi-Fi fingerprinting offers coarse floor cues but is highly sensitive to AP density and enclosure. These results underline the practicality of our ROS barometric service as a drop-in vertical sensor for mobile robots operating in multi-storey buildings where RF infrastructure is sparse and visual features are unreliable.

\section{Conclusion}

This work addressed robust vertical (height and floor) estimation for indoor robots, focusing primarily on a low-cost, differential barometric sensing solution. We presented a ROS-compliant framework that publishes real-time altitude and base reference topics, enabling seamless integration with existing robotic systems. Experimental results in challenging vertical environments—enclosed stairwells and elevators—demonstrated that our barometric method delivers drift-free, sub-meter vertical accuracy and reliable floor recognition, clearly altitude visual SLAM, LiDAR odometry, and Wi-Fi fingerprinting under comparable conditions. 

Future work will focus on further improving vertical accuracy by fusing barometric and IMU data using filtering frameworks (e.g., EKF), and tightly integrating barometric constraints directly into graph-based SLAM back-ends within ROS, thereby enhancing robustness against vertical drift in complex indoor deployments.

\section{Acknowledgment}

The experiments in this work were supported by the Core Facility Platform of Computer Science and Communication, School of Information Science and Technology (SIST), ShanghaiTech University.

\bibliographystyle{IEEEtran}
\bibliography{refs}

\end{document}